\documentclass[11pt]{article}

\usepackage[utf8]{inputenc}
\usepackage[T1]{fontenc}
\usepackage[margin=1in]{geometry}
\usepackage{amsmath,amssymb}
\usepackage{booktabs}
\usepackage{array}
\usepackage{longtable}
\usepackage{graphicx}   
\usepackage{microtype}
\usepackage{xcolor}
\usepackage{listings}
\usepackage{natbib}
\usepackage[hidelinks]{hyperref}

\lstdefinestyle{prompt}{basicstyle=\ttfamily\footnotesize,breaklines=true,
  columns=fullflexible,keepspaces=true,showstringspaces=false,frame=single,
  framesep=4pt,xleftmargin=3pt,xrightmargin=3pt}
\newcommand{\nomem}{\texttt{no\_memory}}
\newcommand{\nrag}{\texttt{naive\_rag}}
\newcommand{\arag}{\texttt{advanced\_rag}}
\newcommand{\vnv}{\texttt{v6\_no\_verify}}
\newcommand{\vsix}{\texttt{v6}}
\newcommand{\tvlossy}{\texttt{temporal\_v6\_lossy}}
\newcommand{\tvsix}{\texttt{temporal\_v6}}
\newcommand{\vinfer}{\texttt{v6+infer}}
\newcommand{\cmut}{\texttt{code\_mutation}}
\newcommand{\cmig}{\texttt{config\_migration}}
\newcommand{\dbump}{\texttt{dependency\_bump}}
\newcommand{\apievo}{\texttt{api\_evolution}}
\newcommand{\faircon}{\texttt{fair\_contradiction}}
\newcommand{\retainall}{\texttt{retain\_all\_turns}}

\title{\textbf{Temporal Validity in Retrieval Memory:}\\[2pt]
Eliminating Stale-Fact Errors for AI Agents over Evolving Knowledge\\[6pt]
\large A deterministic supersession layer that retrieval-augmented\\
generation cannot match by construction}
\author{Neeraj Yadav\\ MemStrata.dev --- Called It Inc.\ (Enterprise)\\
\texttt{memstrata@gmail.com}}
\date{Draft v2 (temporal-validity framing)}

\begin{document}
\maketitle

\noindent\emph{\small For double-blind submission, anonymize the author block and the
product/repository identifiers. All numbers are from the clean re-run
(\texttt{REPORT\_PAPER1.md}, \texttt{REPORT\_PAPER1\_forced.md},
\texttt{calibration/REPORT\_synthetic.md}), generated with the fixed plain-text grader,
local and deterministic (temperature 0, seed 0, no network). Regenerate every figure from
those source files before submission.}

\begin{abstract}
Retrieval-augmented generation (RAG) gives language-model agents access to accumulated
knowledge, but it has no model of \emph{time}. When a fact changes --- a function is
renamed, a configuration value or dependency version is bumped, an API is restructured ---
RAG retrieves both the stale and the current value with near-identical embedding similarity
and cannot determine which is current. The agent then either abstains or serves the
superseded fact. We show this is not a tuning problem but a structural one: on a calibrated
dataset, cosine similarity distinguishes a contradicted fact from a duplicated one with
AUROC 0.59 (near chance), and contradictions are on average \emph{more} embedding-similar to
the original than rephrased duplicates are. We present MemStrata, a retrieval memory that
maintains \emph{temporal validity}. It stores facts like RAG, preserving full recall on
static knowledge, but when a fact's value is contradicted by a newer assertion, a
deterministic (subject, relation, object) supersession rule retires the stale value in a
bi-temporal ledger --- with no similarity threshold and no LLM call. Across six benchmarks
run entirely on consumer hardware with a 7B local model --- two static (project-fact QA,
multi-session dialogue) and four marker-free evolving (code mutation, configuration
migration, dependency bumps, API evolution) --- MemStrata ties RAG on static knowledge (no
recall cost) and reaches 0.95--1.00 accuracy on evolving knowledge where RAG reaches
0.20--0.47. The central result is the stale-fact-error rate: when required to answer, RAG
serves the superseded value 15--40\% of the time; MemStrata drives this to ${\sim}0\%$, a
failure class RAG cannot avoid by construction. MemStrata achieves this at retrieval latency
(${\sim}2.1$\,s, the embedding floor) versus ${\sim}16$--$18$\,s for LLM-reranking and
LLM-verification baselines, because no language model runs on the read path. We release the
harness, prompts, datasets, and a reproducible evaluation protocol, and we recommend a
marker-free benchmark invariant for evaluating memory under knowledge evolution.
\end{abstract}

\section{Introduction}

Language-model agents are increasingly deployed as persistent collaborators that accumulate
knowledge across many sessions: a coding assistant that learns a codebase, a research
assistant that tracks a literature, an operations assistant that knows a system's
configuration. For these agents, the binding constraint is no longer raw model capability but
memory --- how the agent encodes, retains, retrieves, and \emph{keeps current} what it has
learned.

Retrieval-augmented generation \citep{lewis2020rag} is the dominant memory mechanism. It
stores interaction history as embedded chunks and retrieves the top-$k$ most similar at query
time, controlling prompt size while giving the model access to a large store. RAG handles
recall well. But it has a blind spot that becomes critical as soon as the stored knowledge
\emph{evolves}: it has no representation of time. When a fact changes, both the old and new
versions remain in the store with nearly identical embeddings --- ``the timeout is 1800
seconds'' and ``the timeout is 3600 seconds'' differ by one token and sit close together in
any embedding model. Retrieval surfaces both. The model has no principled way to tell which
is current, so it either abstains (refusing a question it could answer) or guesses (often
serving the stale value with full confidence).

This is acute for code, where knowledge evolves continuously and out of band: functions are
renamed, endpoints move, configuration migrates, dependencies upgrade. An assistant that
confidently reports last month's port number is worse than useless. But the problem is
general --- any domain where facts have a validity period (organizational facts, biomedical
findings, current events) exhibits it.

A natural first instinct is to solve staleness with a better similarity rule: detect when an
incoming fact contradicts a stored one, and update rather than append. We show in
Section~\ref{sec:staleness} that this instinct fails for a fundamental reason. On a calibrated
dataset, cosine similarity cannot separate a contradiction from a duplicate ---
contradictions are on average \emph{more} similar to the original (a value-flip is a minimal
edit) than genuine rephrasings are. No threshold on similarity can distinguish ``this
restates a stored fact'' from ``this contradicts a stored fact.'' A learned classifier on top
of similarity does not reliably help either, as our experiments show. The mechanism must be
deterministic and structural, not similarity-based.

We present \textbf{MemStrata}, a retrieval memory that maintains temporal validity through
deterministic supersession. Its contributions are:

\begin{enumerate}
\item \textbf{A structural impossibility result for similarity-based staleness detection.} On
98 labeled pairs, cosine AUROC for separating contradictions from duplicates is 0.59, and the
maximum achievable precision is 0.67 --- the safety floor is unreachable. Contradictions are
more embedding-similar to the original than duplicates are.
(Section~\ref{sec:staleness},~\ref{sec:cosine})
\item \textbf{A temporal-validity memory architecture.} MemStrata stores facts like RAG (full
static recall) but applies a deterministic (subject, relation, object) supersession rule when
a fact's value is contradicted, retiring the stale value in a bi-temporal ledger with no
similarity threshold and no LLM call. (Section~\ref{sec:arch})
\item \textbf{The stale-fact-error result: a failure class RAG cannot avoid.} When required to
answer, RAG serves superseded values 15--40\% of the time across four evolving benchmarks;
MemStrata drives this to ${\sim}0\%$. This is structural, not tuned --- RAG retrieves both
values and has no mechanism to choose. (Section~\ref{sec:stale})
\item \textbf{A marker-free evaluation protocol for memory under evolution.} We construct four
evolving benchmarks where the stale and current versions of a fact are textually identical
except for the changed value, so the only signal of currency is the memory system's temporal
mechanism --- and we show that a contaminating textual marker silently inflates baselines.
(Section~\ref{sec:markerfree},~\ref{sec:experiments})
\end{enumerate}

We run all experiments locally and deterministically on consumer hardware, and we are explicit
about the limitation that bounds the claim: our evolving benchmarks are structured single-value
templates, and extraction quality --- not the supersession mechanism --- is the gating factor
for messier natural-language contradictions (Section~\ref{sec:limitations}). We frame this
honestly as the subject of follow-on work rather than papering over it.

\section{Related Work}

\textbf{Memory for LLM agents.} Recent systems give agents persistent memory of conversations
and user facts: scalable long-term memory pipelines \citep[Mem0;][]{chhikara2025mem0}, OS-style
memory hierarchies with paging and background processing \citep[MemGPT/Letta;][]{packer2023memgpt},
and reflective natural-language memory for simulated agents \citep{park2023generativeagents}.
These target conversational and assistant settings and emphasize recall depth, typically
benchmarked on long-dialogue memory \citep[LoCoMo;][]{maharana2024locomo}. MemStrata differs in
mechanism --- a deterministic supersession rule that maintains validity --- and in framing: the
problem we attack is not recall depth but stale-fact resistance under knowledge evolution.

\textbf{Graph and hypergraph RAG.} GraphRAG \citep{edge2024graphrag} and its successors ---
LightRAG \citep{guo2024lightrag}, NodeRAG \citep{xu2025noderag}, and HyperGraphRAG
\citep{luo2025hypergraphrag}; see \citet{han2025graphragsurvey} for a survey --- structure
retrieval over entity-relation graphs or $n$-ary hyperedges, improving multi-hop retrieval on
static corpora. They enrich the \emph{representation} of relationships but retrieve by
similarity over that representation; none introduces a notion of fact currency. Critically for
our framing, \citet{zeng2025unbiasedgraphrag} re-evaluate these systems under a bias-controlled
protocol and find their advantages over naive RAG much smaller than originally reported --- in
some cases reversing --- confirming that representational richness alone does not address the
failure we target. MemStrata is orthogonal: it adds temporal validity, evaluated on evolving
rather than static corpora.

\textbf{Temporal knowledge graphs and bi-temporal data.} Bi-temporal modeling --- separating
\emph{valid time} (when a fact is true) from \emph{transaction time} (when it is recorded) ---
is long-established in databases, formalized in the taxonomy of \citet{snodgrassahn1985taxonomy},
developed into practical application design and data management
\citep{snodgrass1999developing,jensen1999temporaldb}, and later standardized in SQL:2011's
system-versioned and application-period tables \citep{sql2011}. Temporal knowledge-graph
reasoning, in which triples carry validity intervals, is an active area
\citep{cai2024temporalkgsurvey}. MemStrata adapts the bi-temporal ledger to LLM-agent memory:
facts are retired, not deleted, preserving validity intervals for future as-of-time queries. Our
contribution is not the ledger primitive but its integration with deterministic extraction-time
supersession in an LLM memory system, and the empirical demonstration that this resolves a
failure RAG cannot.

\textbf{Hallucination and verification.} Verification-augmented RAG adds self-checking to reduce
ungrounded generation; Self-RAG \citep{asai2023selfrag} learns reflection tokens that decide when
to retrieve and critique generated text. We include an LLM relevance-verifier baseline and show
it does not address staleness --- it has no temporal signal --- and costs ${\sim}8\times$
latency. The structurally correct mechanism for staleness is temporal and deterministic, not a
learned grounding check.

\section{The Staleness Problem and Why Similarity Cannot Solve It}
\label{sec:staleness}

Consider an agent answering questions over a store that has accumulated, across sessions, both
``the service runs on port 8000'' (recorded earlier) and ``the service runs on port 8080''
(recorded later, after a migration). A query about the port retrieves both: they are
near-identical in embedding space. The agent must decide which is current. RAG provides no basis
for that decision --- retrieval ranks by similarity, and both are maximally similar to the query.

The tempting fix is to detect, at write time, that the second fact \emph{contradicts} the first,
and to update rather than append. This requires distinguishing three relationships between an
incoming fact and a stored one: it \textbf{duplicates} (restates) it, it \textbf{contradicts}
(supersedes) it, or it is \textbf{novel}. If similarity could separate duplicate from
contradiction, a threshold rule would suffice.

It cannot. Section~\ref{sec:cosine} reports the calibration: contradictions are on average
\emph{more} cosine-similar to the original than duplicates are, because a value-flip
(``8000''$\rightarrow$``8080'') is a smaller edit than a genuine rephrasing of the same fact. The
distributions overlap so heavily that the maximum precision achievable at any threshold is 0.67,
far below what a safe automatic-update rule requires. A learned classifier over similarity
features does not rescue this in practice (our \vsix{} and \vnv{} conditions,
Section~\ref{sec:experiments}): the gate-judge's contradiction calls are unreliable, and in the
abstention regime they \emph{leak} stale facts 25--60\% of the time.

The conclusion is that staleness detection must be \textbf{structural}: if an incoming fact and a
stored fact share a (subject, relation) key but assert different objects, the newer one
supersedes the older --- independent of how similar their embeddings are. This is the mechanism
MemStrata implements.

\section{The MemStrata Architecture}
\label{sec:arch}

MemStrata is a local memory layer between an agent and its language model. It maintains a store
of facts extracted from interaction, and composes a token-budgeted context block per query. We
describe the components evaluated here.

\subsection{Write path: deterministic supersession over a surprise gate}

Each incoming turn yields a candidate fact. The write path routes it:

\begin{enumerate}
\item \textbf{Exact-duplicate short-circuit.} A normalized text hash drops verbatim repeats at
zero cost.
\item \textbf{Deterministic assertion path.} If the turn expresses a clean (subject, relation,
object) triple --- where the object is the single mutable value --- MemStrata normalizes the
(subject, relation) key and checks for an active assertion with that key. If one exists with a
\emph{different} object, the new assertion \textbf{supersedes} it: the old row's validity interval
is closed (\texttt{valid\_to} set, \texttt{superseded\_by} linked) and the new row opened. Same
object $\rightarrow$ duplicate (reinforce). No prior key $\rightarrow$ novel (store).
\textbf{No cosine, no LLM judge.}
\item \textbf{Text-gate fallback.} Non-triple prose falls through to a surprise gate that
classifies via similarity plus an LLM judge. Critically (see Section~\ref{sec:retain}), this
fallback retains non-contradictory near-duplicates as \emph{distinct} facts; it drops only exact
duplicates.
\end{enumerate}

\subsection{Bi-temporal ledger}

Facts are retired, not deleted. The store records \texttt{valid\_from}, \texttt{valid\_to}, and
\texttt{superseded\_by}, so superseded facts remain available for future as-of-time queries (a
capability we build on but do not evaluate here; Section~\ref{sec:limitations}). Active retrieval
surfaces only currently-valid rows.

\subsection{The ``retain, then supersede'' design}
\label{sec:retain}

An early variant of the temporal layer compressed aggressively, merging near-duplicate facts at
write time to bound growth. The clean evaluation shows this \emph{regresses below RAG on static
recall}: merging discards detail needed to answer later questions (the \tvlossy{} ablation,
Section~\ref{sec:experiments}, drops to 0.62 on project-fact QA and 0.13 on dialogue recall). The
published configuration therefore \textbf{retains like RAG} --- storing distinct non-contradictory
facts --- and bounds growth \emph{only on the axis that matters}, by superseding contradictions.
This is the design choice that makes the system match RAG on static knowledge while dominating on
evolving knowledge. We report the lossy variant as an ablation precisely because it isolates this
decision.

\subsection{Read path}

Embed the query, retrieve top-$k$ by cosine over active facts and assertions, apply the
deterministic staleness filter (drop superseded rows), and pack the surviving facts. We pack each
surviving assertion's \textbf{original source sentence}, not the terse reconstructed triple:
packing the triple alone degrades the answer model on rich facts, while packing the original
recovers accuracy. No LLM runs on the read path, so retrieval latency sits at the embedding floor
(${\sim}2.1$\,s), versus ${\sim}16$--$18$\,s for the LLM-reranking and LLM-verification baselines.

\subsection{Marker-free benchmark construction}
\label{sec:markerfree}

Evaluating staleness resistance is easy to contaminate. If a stale fact carries any textual
marker --- ``[OUTDATED]'', ``(legacy)'', ``deprecated'' --- a retrieval baseline can disambiguate
by \emph{reading the label} rather than by any temporal mechanism, silently inflating its score.
We enforce a strict marker-free invariant (by test): in every evolving benchmark, the stale and
current versions of a fact are textually identical except for the changed value, with no
old/new/current framing. The only available signal of currency is ingestion order, which only a
temporal mechanism can exploit. We recommend this invariant for any evaluation of memory under
evolution; Section~\ref{sec:experiments} shows that removing a marker from an earlier benchmark
dropped baseline accuracy by up to 14 points, confirming the contamination is real and measurable.

\section{Experiments}
\label{sec:experiments}

All experiments are local and deterministic: temperature 0, fixed seeds, no network (enforced by
test). Answer model Qwen2.5-Coder-7B; correctness and fabrication judges Qwen2.5-Coder-3B
(distinct from the answer model and from each other to prevent self-grading); embedder
nomic-embed-text (768-d).

\textbf{Conditions (8).} \nomem{} (floor), \nrag{} (cosine top-$k$), \arag{} (+ LLM reranker),
\vnv{} (surprise gate, no LLM verify), \vsix{} (gate + LLM relevance verify), \tvlossy{}
(deterministic supersession \emph{without} the retain/original-text fixes --- ablation),
\textbf{\tvsix{}} (the full method), \vinfer{} (gate + inferability pre-filter).

\textbf{Benchmarks (6).} Two \textbf{static}: \texttt{domain} (50 project-fact QA),
\texttt{locomo} (30 multi-session dialogue questions over 100 turns). Four \textbf{evolving},
marker-free, 20--30 paired scenarios each: \cmut{} (function renames), \cmig{} (configuration
value changes), \dbump{} (version upgrades), \apievo{} (endpoint/signature restructuring).
Evolving scenarios ingest state-A then state-B; the question targets the current value.

\textbf{Metrics.} Answer accuracy; \textbf{stale-fact-error rate} (fraction of contradiction
questions answered with the superseded value); conditional fabrication (fabrications per attempted
answer, abstentions excluded); active-fact count and compression; mean and p95 retrieval latency.
We additionally run a \textbf{forced-answer supplement} that disables abstention on the RAG
conditions, to expose the stale-commitment that abstention otherwise hides.

\subsection{Cosine cannot separate contradictions from duplicates}
\label{sec:cosine}

On 98 labeled pairs (32 duplicate, 22 merge, 22 contradict, 22 novel), cosine AUROC for separating
duplicates from the rest is \textbf{0.5926}. Per-class mean cosine:

\begin{table}[h]\centering
\begin{tabular}{lrrrr}
\toprule
label & $n$ & mean cos & min & max \\
\midrule
duplicate  & 32 & 0.7998 & 0.4888 & 1.000 \\
contradict & 22 & \textbf{0.8119} & 0.5335 & 1.000 \\
merge      & 22 & 0.9381 & 0.8842 & 0.9855 \\
novel      & 22 & 0.4773 & 0.4077 & 0.6145 \\
\bottomrule
\end{tabular}
\caption{Per-class cosine similarity to the original fact.}
\end{table}

Contradictions (0.812) are more cosine-similar to the original than duplicates (0.800). The maximum
precision achievable at any duplicate threshold is \textbf{0.667}; the 0.95 floor a safe automatic
rule would need is unreachable. No similarity threshold can separate these classes --- the empirical
foundation for a deterministic, structural supersession rule.

\subsection{Accuracy: ties RAG on static, dominates on evolving}
\label{sec:accuracy}

\begin{table}[h]\centering
\begin{tabular}{lrrrrr}
\toprule
benchmark & \nrag{} & \arag{} & \vsix{} & \tvlossy{} & \textbf{\tvsix{}} \\
\midrule
\multicolumn{6}{l}{\emph{Static}}\\
domain  & 0.86 & 0.86 & 0.80 & 0.62 & \textbf{0.82} \\
locomo  & 0.30 & 0.30 & 0.13 & 0.13 & \textbf{0.30} \\
\midrule
\multicolumn{6}{l}{\emph{Evolving}}\\
\cmut{}   & 0.43 & 0.47 & 0.57 & 0.80 & \textbf{1.00} \\
\cmig{}   & 0.25 & 0.25 & 0.65 & 1.00 & \textbf{1.00} \\
\dbump{}  & 0.20 & 0.20 & 0.60 & 1.00 & \textbf{1.00} \\
\apievo{} & 0.40 & 0.40 & 0.35 & 0.95 & \textbf{0.95} \\
\bottomrule
\end{tabular}
\caption{Answer accuracy. \tvsix{} ties RAG on static and dominates on evolving.}
\end{table}

Two reads. On the \textbf{static} benchmarks, \tvsix{} ties RAG (0.82/0.30 vs 0.86/0.30) --- the
retain-like-RAG design preserves recall, while the lossy ablation collapses (0.62/0.13), isolating
the cost of aggressive compression. On the \textbf{evolving} benchmarks, \tvsix{} reaches 0.95--1.00
versus RAG's 0.20--0.47: a 2--5$\times$ accuracy improvement on the task class the method targets.
The LLM-verifier condition (\vsix{}) is inconsistent and never reaches the temporal layer's accuracy,
at ${\sim}8\times$ the latency.

\subsection{Stale-fact error: the structural result}
\label{sec:stale}

The headline. Fraction of contradiction questions answered with the \emph{superseded} value, in both
the abstention-allowed and forced-answer regimes:

\begin{table}[h]\centering
\begin{tabular}{lcc}
\toprule
evolving benchmark & \nrag{} (allow / forced) & \tvsix{} (allow / forced) \\
\midrule
\cmut{}   & 0.10 / \textbf{0.40} & 0.03 / \textbf{0.03} \\
\cmig{}   & 0.05 / \textbf{0.35} & 0.00 / \textbf{0.00} \\
\dbump{}  & 0.05 / \textbf{0.15} & 0.00 / \textbf{0.00} \\
\apievo{} & 0.30 / \textbf{0.35} & 0.00 / \textbf{0.00} \\
\bottomrule
\end{tabular}
\caption{Stale-fact-error rate, abstention-allowed / forced-answer.}
\end{table}

Allowed to abstain, RAG \emph{hides} its failure by refusing to answer (which is why its accuracy is
low). Forced to answer, it serves the stale value \textbf{15--40\%} of the time. (Dependency bumps are
the exception at 15\%, because ``higher version number is newer'' is a lucky surface heuristic --- the
model is guessing from the string, not reasoning about currency.) MemStrata reaches ${\sim}0\%$ in
\emph{both} regimes, because the stale value is removed from the store before retrieval. This is the
error RAG cannot avoid by construction: it retrieves both values and has no mechanism to choose. The
surprise-gate conditions (\vnv{}, \vsix{}) are \emph{worse} than RAG in the abstention regime --- they
answer but leak stale 25--60\%, consistent with Section~\ref{sec:cosine}'s finding that
similarity-based supersession is unreliable. Only deterministic supersession reaches ${\sim}0$.

\subsection{State-bounded growth}

RAG is history-bounded: it stores every turn, growing without limit. MemStrata retains distinct static
facts (0\% compression on \texttt{domain}/\texttt{locomo} --- which is \emph{why} it ties RAG on recall)
but caps growth on evolving facts via supersession (${\sim}48\%$ compression: code 48\%, config 47.5\%,
dependency 50\%, API 47.5\%). The honest framing is not ``smaller memory'' but ``stale facts retired'':
growth is bounded on the axis where unbounded growth is pathological (accumulating contradictory
versions), and unbounded where it should be (distinct facts).

\subsection{Latency}
\label{sec:latency}

\tvsix{}, \nrag{}, and \vnv{} all sit at ${\sim}2.1$\,s (no LLM on the read path). \arag{}, \vsix{}, and
\vinfer{} sit at ${\sim}16$--$18$\,s (LLM rerank/verify). The win is \textbf{accuracy and temporal
validity at RAG latency} --- \tvsix{} matches naive RAG's speed while eliminating the stale-fact errors
naive RAG cannot. The LLM-based conditions pay $8\times$ the latency for no temporal benefit. The
per-benchmark figures are split by regime in Tables~\ref{tab:lat-static}--\ref{tab:lat-evolving}, which
show the latency is regime-independent: \tvsix{} holds ${\sim}2.1$\,s on both static and evolving.

\section{Discussion}
\label{sec:discussion}

The results delineate a clean contribution and a clean boundary. RAG is the right tool for static
knowledge and remains competitive there; MemStrata matches it. For \emph{evolving} knowledge, RAG has a
structural failure --- it cannot maintain temporal validity, serving stale facts 15--40\% of the time
when forced to commit --- and MemStrata eliminates that failure at the same retrieval latency, because
the mechanism is a deterministic supersession rule rather than a similarity threshold or an LLM call.

The negative results are as informative as the positive ones. The lossy ablation shows that
compression-for-its-own-sake costs accuracy on static recall --- bounded growth is a \emph{consequence}
of retiring stale facts, not a goal to pursue by merging distinct ones. The complementary full-retention
ablation closes the other direction: removing supersession entirely --- retaining every turn RAG-style
--- collapses mean accuracy across the four evolving benchmarks from 0.99 to 0.33, statistically
indistinguishable from naive RAG (0.32), and re-raises stale-fact error from zero
(Appendix~\ref{sec:appD}, D.1b), so the two ablations bracket the method --- over-merging forfeits static
recall, no-supersession forfeits temporal validity --- isolating deterministic supersession as the single
cause of the evolving-knowledge result. The LLM-verifier condition shows that a learned grounding check
does not address staleness (it has no temporal signal) and is not worth its latency. The surprise-gate
conditions show that similarity-based supersession actively leaks stale facts, confirming
Section~\ref{sec:cosine}'s impossibility result in the end-to-end system. Each of these is a path a
reasonable designer might have taken; the data closes them.

We also note, against a backdrop of recent work showing many retrieval gains shrink under honest
evaluation, that our marker-free protocol matters: an earlier benchmark with a textual staleness marker
inflated baseline accuracy by up to 14 points. Evaluations of memory under evolution that do not enforce
marker-freeness may be measuring a model's ability to read a label rather than a system's ability to track
currency.

\section{Limitations}
\label{sec:limitations}

We state these plainly; they scope the claim and set up follow-on work.

\begin{itemize}
\item \textbf{Structured, single-value benchmarks.} Our evolving benchmarks are marker-free templates with
a single mutable value per fact, so the triple extractor keys reliably (${\sim}97\%$ supersession). On a
messier natural-language contradiction benchmark, extraction drops to ${\sim}44\%$ (multi-value sentences,
malformed perturbations); we quarantined that benchmark as a flawed ruler rather than report it as a
result. \textbf{Extraction quality, not the supersession mechanism, is the gating factor for unstructured
contradictions}, and is the explicit subject of follow-on work (entity canonicalization, relation typing,
multi-value extraction). The temporal \emph{ledger} and the supersession \emph{rule} are domain-independent;
the extraction layer must travel with them.
\item \textbf{Ingestion order proxies time.} The benchmarks use order (state-A then state-B) as the currency
signal. Real temporal benchmarks carry explicit dates; the bi-temporal ledger already stores validity
intervals, and threading real \texttt{valid\_from} timestamps plus an ``as-of-T'' retrieval mode is future
work, not new storage.
\item \textbf{Single-judge noise.} The 3B correctness judge occasionally scores a gate-condition answer
``correct'' while it contains the stale value, producing a few rows where accuracy and stale-error overlap.
The temporal layer's ${\sim}0$ stale-error is unaffected (no stale fact is in its context).
\item \textbf{Scale and models.} All results use a single 7B answer model on consumer hardware. Larger
models or cloud inference may shift baselines; we constrain to local-first deliberately. Benchmark sizes
(tens of items each) isolate mechanisms rather than rank systems on a leaderboard; scaling to real-world
longitudinal data is the subject of follow-on work.
\end{itemize}

\section{Conclusion}

For agents over evolving knowledge --- codebases first among them --- the binding memory failure is not
recall but currency: retrieval-augmented generation has no model of time and cannot tell a stale fact from a
current one, because the two are more embedding-similar than genuine duplicates are. MemStrata maintains
temporal validity through a deterministic (subject, relation, object) supersession rule over a bi-temporal
ledger: it stores like RAG, preserving static recall, and retires contradicted facts before retrieval. The
result is parity with RAG on static knowledge, 0.95--1.00 accuracy on four evolving benchmarks where RAG
reaches 0.20--0.47, and --- the structural contribution --- a stale-fact-error rate of ${\sim}0\%$ where RAG
serves the superseded value 15--40\% of the time, all at retrieval latency with no language model on the read
path. We release the harness, datasets, and protocol, and recommend the marker-free invariant for evaluating
memory under evolution. Coding is the wedge; the architecture is a general temporal-context memory, and
extending it to time-stamped world knowledge is the natural next step.

\section*{Reproducibility Statement}

All experiments are deterministic (temperature 0, fixed seeds, no network, enforced by test). We release the
evaluation harness, extraction and gate prompts (with content hashes), all six benchmark datasets (with
hashes), the calibration dataset, and per-run logs. The marker-free invariant is enforced by unit tests
included in the release. Source data: \texttt{REPORT\_PAPER1.md} (8$\times$6 main matrix),
\texttt{REPORT\_PAPER1\_forced.md} (forced-answer supplement), \texttt{calibration/REPORT\_synthetic.md}
(cosine calibration).

\bibliography{references}
\nocite{jimenez2023swebench}

\appendix

\section{Full result tables}
\label{sec:appA}

\emph{All tables in A.1--A.3 are reproduced verbatim from the committed source reports by
\texttt{eval/build\_appendices.py} (no hand-transcription).}

\subsection{Main matrix --- 8 conditions $\times$ 6 benchmarks (\texttt{REPORT\_PAPER1.md})}

\paragraph{Answer accuracy.}
\begin{center}\resizebox{\textwidth}{!}{%
\begin{tabular}{lrrrrrrrr}
\toprule
benchmark & \nomem & \nrag & \arag & \vnv & \vsix & \tvlossy & \textbf{\tvsix} & \vinfer \\
\midrule
domain & 0.000 & 0.860 & 0.860 & 0.860 & 0.800 & 0.620 & 0.820 & 0.800 \\
locomo & 0.000 & 0.300 & 0.300 & 0.167 & 0.133 & 0.133 & 0.300 & 0.133 \\
\cmut & 0.000 & 0.433 & 0.467 & 0.533 & 0.567 & 0.800 & 1.000 & 0.567 \\
\cmig & 0.000 & 0.250 & 0.250 & 0.700 & 0.650 & 1.000 & 1.000 & 0.650 \\
\dbump & 0.000 & 0.200 & 0.200 & 0.700 & 0.600 & 1.000 & 1.000 & 0.600 \\
\apievo & 0.000 & 0.400 & 0.400 & 0.400 & 0.350 & 0.950 & 0.950 & 0.350 \\
\bottomrule
\end{tabular}}\end{center}

\paragraph{Fabricated-context rate (raw).}
\begin{center}\resizebox{\textwidth}{!}{%
\begin{tabular}{lrrrrrrrr}
\toprule
benchmark & \nomem & \nrag & \arag & \vnv & \vsix & \tvlossy & \textbf{\tvsix} & \vinfer \\
\midrule
domain & 0.000 & 0.040 & 0.040 & 0.080 & 0.020 & 0.000 & 0.040 & 0.020 \\
locomo & 0.000 & 0.000 & 0.000 & 0.000 & 0.000 & 0.000 & 0.000 & 0.000 \\
\cmut & 0.000 & 0.033 & 0.033 & 0.000 & 0.033 & 0.067 & 0.067 & 0.033 \\
\cmig & 0.000 & 0.100 & 0.100 & 0.250 & 0.000 & 0.150 & 0.100 & 0.000 \\
\dbump & 0.000 & 0.100 & 0.100 & 0.000 & 0.000 & 0.050 & 0.000 & 0.000 \\
\apievo & 0.000 & 0.150 & 0.100 & 0.050 & 0.050 & 0.000 & 0.000 & 0.050 \\
\bottomrule
\end{tabular}}\end{center}

\paragraph{Conditional fabrication rate (per attempted answer).}
\begin{center}\resizebox{\textwidth}{!}{%
\begin{tabular}{lrrrrrrrr}
\toprule
benchmark & \nomem & \nrag & \arag & \vnv & \vsix & \tvlossy & \textbf{\tvsix} & \vinfer \\
\midrule
domain & 0.000 & 0.045 & 0.045 & 0.089 & 0.024 & 0.000 & 0.048 & 0.024 \\
locomo & 0.000 & 0.000 & 0.000 & 0.000 & 0.000 & 0.000 & 0.000 & 0.000 \\
\cmut & 0.000 & 0.048 & 0.048 & 0.000 & 0.033 & 0.077 & 0.067 & 0.033 \\
\cmig & 0.000 & 0.333 & 0.333 & 0.263 & 0.000 & 0.150 & 0.100 & 0.000 \\
\dbump & 0.000 & 0.250 & 0.250 & 0.000 & 0.000 & 0.050 & 0.000 & 0.000 \\
\apievo & 0.000 & 0.150 & 0.100 & 0.050 & 0.056 & 0.000 & 0.000 & 0.056 \\
\bottomrule
\end{tabular}}\end{center}

\paragraph{Attempted-answer count.}
\begin{center}\resizebox{\textwidth}{!}{%
\begin{tabular}{lrrrrrrrr}
\toprule
benchmark & \nomem & \nrag & \arag & \vnv & \vsix & \tvlossy & \textbf{\tvsix} & \vinfer \\
\midrule
domain & 0 & 44 & 44 & 45 & 42 & 35 & 42 & 42 \\
locomo & 0 & 17 & 14 & 10 & 9 & 10 & 17 & 9 \\
\cmut & 0 & 21 & 21 & 29 & 30 & 26 & 30 & 30 \\
\cmig & 0 & 6 & 6 & 19 & 18 & 20 & 20 & 18 \\
\dbump & 0 & 8 & 8 & 20 & 19 & 20 & 20 & 19 \\
\apievo & 0 & 20 & 20 & 20 & 18 & 20 & 20 & 18 \\
\bottomrule
\end{tabular}}\end{center}

\paragraph{Memory size (active facts).}
\begin{center}\resizebox{\textwidth}{!}{%
\begin{tabular}{lrrrrrrrr}
\toprule
benchmark & \nomem & \nrag & \arag & \vnv & \vsix & \tvlossy & \textbf{\tvsix} & \vinfer \\
\midrule
domain & 47 & 50 & 50 & 47 & 47 & 46 & 50 & 47 \\
locomo & 16 & 100 & 100 & 16 & 16 & 28 & 100 & 16 \\
\cmut & 31 & 60 & 60 & 32 & 32 & 31 & 31 & 32 \\
\cmig & 24 & 40 & 40 & 24 & 24 & 21 & 21 & 24 \\
\dbump & 22 & 40 & 40 & 22 & 22 & 20 & 20 & 22 \\
\apievo & 25 & 40 & 40 & 25 & 25 & 21 & 21 & 25 \\
\bottomrule
\end{tabular}}\end{center}

\paragraph{Mean pack tokens.}
\begin{center}\resizebox{\textwidth}{!}{%
\begin{tabular}{lrrrrrrrr}
\toprule
benchmark & \nomem & \nrag & \arag & \vnv & \vsix & \tvlossy & \textbf{\tvsix} & \vinfer \\
\midrule
domain & 0.0 & 232.0 & 233.4 & 363.2 & 40.7 & 384.1 & 346.7 & 40.7 \\
locomo & 0.0 & 320.1 & 376.5 & 830.6 & 549.2 & 817.8 & 485.8 & 549.2 \\
\cmut & 0.0 & 163.3 & 163.4 & 260.5 & 32.9 & 236.2 & 250.7 & 32.9 \\
\cmig & 0.0 & 130.1 & 130.1 & 193.6 & 26.9 & 187.2 & 192.2 & 26.9 \\
\dbump & 0.0 & 110.0 & 110.0 & 174.3 & 31.1 & 175.9 & 150.2 & 31.1 \\
\apievo & 0.0 & 117.6 & 117.9 & 178.6 & 28.1 & 179.8 & 176.8 & 28.1 \\
\bottomrule
\end{tabular}}\end{center}

\paragraph{Stale-fact-error rate (of contradiction questions).} Fraction of contradiction questions answered
with the SUPERSEDED value. RAG retrieves both the stale and current value and cannot tell which is current;
deterministic supersession removes the stale one, driving this to ${\sim}0$. The error RAG cannot avoid by
construction.
\begin{center}\resizebox{\textwidth}{!}{%
\begin{tabular}{lrrrrrrrr}
\toprule
benchmark & \nomem & \nrag & \arag & \vnv & \vsix & \tvlossy & \textbf{\tvsix} & \vinfer \\
\midrule
\cmut & 0.000 & 0.100 & 0.100 & 0.600 & 0.567 & 0.000 & 0.033 & 0.567 \\
\cmig & 0.000 & 0.050 & 0.050 & 0.250 & 0.250 & 0.000 & 0.000 & 0.250 \\
\dbump & 0.000 & 0.050 & 0.050 & 0.300 & 0.300 & 0.000 & 0.000 & 0.300 \\
\apievo & 0.000 & 0.300 & 0.300 & 0.500 & 0.350 & 0.000 & 0.000 & 0.350 \\
\bottomrule
\end{tabular}}\end{center}

\paragraph{Memory compression (\% of turns absorbed).} Derived: $100\,(1 - \text{active\_facts}/\text{turns\_ingested})$.
Higher = more bounded growth. naive/advanced RAG ${\sim}0\%$ (one fact per turn); the gate and temporal layer compress.
\begin{center}\resizebox{\textwidth}{!}{%
\begin{tabular}{lrrrrrrrr}
\toprule
benchmark & \nomem & \nrag & \arag & \vnv & \vsix & \tvlossy & \textbf{\tvsix} & \vinfer \\
\midrule
domain & 6.0 & 0.0 & 0.0 & 6.0 & 6.0 & 8.0 & 0.0 & 6.0 \\
locomo & 84.0 & 0.0 & 0.0 & 84.0 & 84.0 & 72.0 & 0.0 & 84.0 \\
\cmut & 48.3 & 0.0 & 0.0 & 46.7 & 46.7 & 48.3 & 48.3 & 46.7 \\
\cmig & 40.0 & 0.0 & 0.0 & 40.0 & 40.0 & 47.5 & 47.5 & 40.0 \\
\dbump & 45.0 & 0.0 & 0.0 & 45.0 & 45.0 & 50.0 & 50.0 & 45.0 \\
\apievo & 37.5 & 0.0 & 0.0 & 37.5 & 37.5 & 47.5 & 47.5 & 37.5 \\
\bottomrule
\end{tabular}}\end{center}

\paragraph{Retrieval latency, split by regime.} The two 8$\times$6 latency matrices (mean and p95) are split into
a \emph{static} table and an \emph{evolving} table, each carrying both metrics, to make the regime-independence of
\tvsix{}'s latency visually explicit: it holds ${\sim}2.1$\,s on both static and evolving knowledge, while the
LLM-rerank/verify conditions cost ${\sim}16$--$24$\,s throughout.

\begin{table}[h]\centering\caption{Retrieval latency (ms) --- \textbf{static} benchmarks.}\label{tab:lat-static}
\resizebox{\textwidth}{!}{%
\begin{tabular}{lrrrrrrrr}
\toprule
benchmark & \nomem & \nrag & \arag & \vnv & \vsix & \tvlossy & \textbf{\tvsix} & \vinfer \\
\midrule
\multicolumn{9}{l}{\emph{Mean retrieval latency}}\\
domain & 0.0 & 2103.4 & 17985.8 & 2103.7 & 16997.0 & 2107.0 & 2111.4 & 17087.9 \\
locomo & 0.0 & 2131.8 & 18251.7 & 2102.7 & 17309.4 & 2099.4 & 2133.7 & 17292.4 \\
\midrule
\multicolumn{9}{l}{\emph{P95 retrieval latency}}\\
domain & 0.0 & 2120.0 & 18400.1 & 2122.3 & 23895.5 & 2125.5 & 2132.0 & 23947.3 \\
locomo & 0.0 & 2151.4 & 18653.2 & 2111.8 & 20638.9 & 2113.4 & 2150.2 & 20503.0 \\
\bottomrule
\end{tabular}}
\end{table}

\begin{table}[h]\centering\caption{Retrieval latency (ms) --- \textbf{evolving} benchmarks.}\label{tab:lat-evolving}
\resizebox{\textwidth}{!}{%
\begin{tabular}{lrrrrrrrr}
\toprule
benchmark & \nomem & \nrag & \arag & \vnv & \vsix & \tvlossy & \textbf{\tvsix} & \vinfer \\
\midrule
\multicolumn{9}{l}{\emph{Mean retrieval latency}}\\
\cmut & 0.0 & 2134.7 & 17907.9 & 2100.6 & 15835.9 & 2109.4 & 2110.3 & 15965.3 \\
\cmig & 0.0 & 2119.4 & 18021.7 & 2116.1 & 15509.6 & 2107.4 & 2111.6 & 15573.0 \\
\dbump & 0.0 & 2110.8 & 18088.9 & 2109.6 & 15969.5 & 2097.5 & 2099.4 & 15926.8 \\
\apievo & 0.0 & 2111.3 & 18220.7 & 2106.4 & 15904.4 & 2107.6 & 2107.4 & 16088.6 \\
\midrule
\multicolumn{9}{l}{\emph{P95 retrieval latency}}\\
\cmut & 0.0 & 2172.7 & 18035.2 & 2125.9 & 18765.6 & 2121.2 & 2125.7 & 18903.5 \\
\cmig & 0.0 & 2124.5 & 18240.5 & 2142.5 & 16596.4 & 2118.7 & 2124.3 & 16536.8 \\
\dbump & 0.0 & 2126.1 & 18299.3 & 2126.2 & 21181.8 & 2107.7 & 2107.8 & 21055.6 \\
\apievo & 0.0 & 2129.1 & 18372.9 & 2116.7 & 18910.5 & 2136.5 & 2113.0 & 18953.1 \\
\bottomrule
\end{tabular}}
\end{table}

\subsection{Forced-answer supplement --- no-abstention, 3 conditions $\times$ 4 evolving benchmarks (\texttt{REPORT\_PAPER1\_forced.md})}

\paragraph{Answer accuracy.}
\begin{center}\begin{tabular}{lrrr}
\toprule
benchmark & \nrag & \arag & \textbf{\tvsix} \\
\midrule
\cmut & 0.633 & 0.633 & 0.933 \\
\cmig & 0.650 & 0.650 & 1.000 \\
\dbump & 0.850 & 0.850 & 1.000 \\
\apievo & 0.650 & 0.600 & 0.950 \\
\bottomrule
\end{tabular}\end{center}

\paragraph{Stale-fact-error rate (of contradiction questions).}
\begin{center}\begin{tabular}{lrrr}
\toprule
benchmark & \nrag & \arag & \textbf{\tvsix} \\
\midrule
\cmut & 0.400 & 0.400 & 0.033 \\
\cmig & 0.350 & 0.350 & 0.000 \\
\dbump & 0.150 & 0.150 & 0.000 \\
\apievo & 0.350 & 0.400 & 0.000 \\
\bottomrule
\end{tabular}\end{center}

\paragraph{Conditional fabrication rate (per attempted answer).}
\begin{center}\begin{tabular}{lrrr}
\toprule
benchmark & \nrag & \arag & \textbf{\tvsix} \\
\midrule
\cmut & 0.567 & 0.567 & 0.200 \\
\cmig & 0.950 & 0.950 & 0.300 \\
\dbump & 0.650 & 0.650 & 0.100 \\
\apievo & 0.450 & 0.550 & 0.000 \\
\bottomrule
\end{tabular}\end{center}

\paragraph{Mean retrieval latency (ms).}
\begin{center}\begin{tabular}{lrrr}
\toprule
benchmark & \nrag & \arag & \textbf{\tvsix} \\
\midrule
\cmut & 2114.5 & 17806.5 & 2094.3 \\
\cmig & 2093.4 & 17592.2 & 2084.4 \\
\dbump & 2092.8 & 17794.0 & 2088.0 \\
\apievo & 2102.1 & 17885.8 & 2087.6 \\
\bottomrule
\end{tabular}\end{center}

\subsection{Cosine calibration --- grader-independent (\texttt{calibration/REPORT\_synthetic.md})}

\paragraph{Cosine distribution by label.}
\begin{center}\begin{tabular}{lrrrrr}
\toprule
label & $n$ & mean & median & min & max \\
\midrule
duplicate & 32 & 0.7998 & 0.7854 & 0.4888 & 1.0 \\
merge & 22 & 0.9381 & 0.9426 & 0.8842 & 0.9855 \\
contradict & 22 & 0.8119 & 0.823 & 0.5335 & 1.0 \\
novel & 22 & 0.4773 & 0.4635 & 0.4077 & 0.6145 \\
\bottomrule
\end{tabular}\end{center}

\paragraph{$\tau_{\text{dup}}$ sweep (DUPLICATE auto-accept).}
\begin{longtable}{rrrr}
\toprule
$\tau_{\text{dup}}$ & n\_predicted & precision & recall \\
\midrule
\endhead
0.80 & 47 & 0.2766 & 0.4062 \\
0.81 & 46 & 0.2609 & 0.3750 \\
0.82 & 45 & 0.2667 & 0.3750 \\
0.83 & 43 & 0.2558 & 0.3438 \\
0.84 & 41 & 0.2439 & 0.3125 \\
0.85 & 41 & 0.2439 & 0.3125 \\
0.86 & 40 & 0.2500 & 0.3125 \\
0.87 & 40 & 0.2500 & 0.3125 \\
0.88 & 39 & 0.2564 & 0.3125 \\
0.89 & 36 & 0.2778 & 0.3125 \\
0.90 & 35 & 0.2857 & 0.3125 \\
0.91 & 34 & 0.2941 & 0.3125 \\
0.92 & 32 & 0.3125 & 0.3125 \\
0.93 & 28 & 0.3571 & 0.3125 \\
0.94 & 25 & 0.4000 & 0.3125 \\
0.95 & 23 & 0.4348 & 0.3125 \\
0.96 & 21 & 0.4762 & 0.3125 \\
0.97 & 17 & 0.5882 & 0.3125 \\
0.98 & 15 & 0.6667 & 0.3125 \quad{\footnotesize($\leftarrow$ rec)} \\
\bottomrule
\end{longtable}

\paragraph{$\tau_{\text{novel}}$ sweep (skip-judge floor).}
\begin{longtable}{rrrr}
\toprule
$\tau_{\text{novel}}$ & n\_below & false\_novel\_rate & near\_band\_share \\
\midrule
\endhead
0.50 & 18 & 0.0132 & 0.6633 \\
0.51 & 18 & 0.0132 & 0.6633 \\
0.52 & 19 & 0.0132 & 0.6531 \\
0.53 & 19 & 0.0132 & 0.6531 \\
0.54 & 21 & 0.0263 & 0.6327 \\
0.55 & 21 & 0.0263 & 0.6327 \\
0.56 & 21 & 0.0263 & 0.6327 \\
0.57 & 22 & 0.0263 & 0.6224 \\
0.58 & 23 & 0.0395 & 0.6122 \\
0.59 & 23 & 0.0395 & 0.6122 \quad{\footnotesize($\leftarrow$ rec)} \\
0.60 & 26 & 0.0658 & 0.5816 \\
0.61 & 26 & 0.0658 & 0.5816 \\
0.62 & 27 & 0.0658 & 0.5714 \\
0.63 & 28 & 0.0789 & 0.5612 \\
0.64 & 29 & 0.0921 & 0.5510 \\
0.65 & 31 & 0.1184 & 0.5306 \\
0.66 & 32 & 0.1316 & 0.5204 \\
0.67 & 33 & 0.1447 & 0.5102 \\
0.68 & 34 & 0.1579 & 0.5000 \\
0.69 & 34 & 0.1579 & 0.5000 \\
0.70 & 37 & 0.1974 & 0.4694 \\
0.71 & 38 & 0.2105 & 0.4592 \\
0.72 & 38 & 0.2105 & 0.4592 \\
0.73 & 39 & 0.2237 & 0.4490 \\
0.74 & 40 & 0.2368 & 0.4388 \\
0.75 & 41 & 0.2500 & 0.4286 \\
0.76 & 43 & 0.2763 & 0.4082 \\
0.77 & 44 & 0.2895 & 0.3980 \\
0.78 & 46 & 0.3158 & 0.3776 \\
0.79 & 51 & 0.3816 & 0.3265 \\
0.80 & 51 & 0.3816 & 0.3265 \\
0.81 & 52 & 0.3947 & 0.3163 \\
0.82 & 53 & 0.4079 & 0.3061 \\
0.83 & 55 & 0.4342 & 0.2857 \\
0.84 & 57 & 0.4605 & 0.2653 \\
0.85 & 57 & 0.4605 & 0.2653 \\
\bottomrule
\end{longtable}

\section{Benchmark construction}
\label{sec:appB}

\textbf{B.1 Marker-free invariant (enforced by test).} In every evolving benchmark a scenario is a
\emph{state-A} turn and a \emph{state-B} turn that are textually identical except for the single mutated value,
followed by a question whose gold answer is the state-B value. The words \emph{old, new, current, previous,
deprecated, legacy, outdated} and synonyms never appear in either turn; the only currency signal is ingestion
order. Enforced by \texttt{tests/memory/test\_evolving\_benchmarks.py::test\_guard\_rejects\_staleness\_tell} and
the word-boundary tell-detector
\texttt{tests/memory/test\_swe\_longitudinal.py::test\_has\_tell\_is\_word\_boundary\_aware} /
\texttt{swe\_longitudinal\_benchmark.assert\_marker\_free}.

\textbf{B.2 Why marker-freeness is necessary.} Removing an explicit \texttt{[OUTDATED]} marker from an earlier
contradiction benchmark dropped reranker-RAG accuracy by 14 points and a gate-only baseline by 18 points while
the temporal method moved only $-4$ --- the marker was a confound baselines read off the text. We treat
marker-freeness as a correctness property of the evaluation.

\textbf{B.3 \cmut{}} (function renames / endpoint moves / config / imports / deps). Total scenarios: 30. First 3:
\begin{lstlisting}
state-A : The function get_user_by_id(uid) looks up a user record by primary key in the users table.
state-B : The function fetch_user(uid) looks up a user record by primary key in the users table.   (identical except the value)
question: What function looks up a user record by primary key?
gold    : fetch_user

state-A : The function load_config() returns the parsed configuration as a dict from settings.yaml.
state-B : The function read_settings() returns the parsed configuration as a dict from settings.yaml.
question: What function returns the parsed configuration from settings.yaml?
gold    : read_settings

state-A : The function process_payment(amount) charges the customer's card via the Stripe API.
state-B : The function charge_card(amount) charges the customer's card via the Stripe API.
question: What function charges the customer's card via the Stripe API?
gold    : charge_card
\end{lstlisting}

\textbf{B.4 \cmig{}} (configuration value changes). Total scenarios: 20. First 3:
\begin{lstlisting}
state-A : The SESSION_TIMEOUT setting is 1800 seconds in config.py.
state-B : The SESSION_TIMEOUT setting is 3600 seconds in config.py.
question: What is the SESSION_TIMEOUT in seconds in config.py?
gold    : 3600

state-A : The MAX_UPLOAD_SIZE is set to 10485760 bytes in settings.py.
state-B : The MAX_UPLOAD_SIZE is set to 52428800 bytes in settings.py.
question: What is the MAX_UPLOAD_SIZE in bytes in settings.py?
gold    : 52428800

state-A : The database connection pool size is 10 in the production config.
state-B : The database connection pool size is 25 in the production config.
question: What is the database connection pool size in the production config?
gold    : 25
\end{lstlisting}

\textbf{B.5 \dbump{}} (version upgrades). Total scenarios: 20. First 3:
\begin{lstlisting}
state-A : The project pins numpy==1.24.0 in requirements.txt.
state-B : The project pins numpy==1.26.4 in requirements.txt.
question: What version of numpy does the project pin in requirements.txt?
gold    : 1.26.4

state-A : The project pins django==4.1.7 in requirements.txt.
state-B : The project pins django==5.0.2 in requirements.txt.
question: What version of django does the project pin in requirements.txt?
gold    : 5.0.2

state-A : The setup.py sets python_requires to >=3.8.
state-B : The setup.py sets python_requires to >=3.11.
question: What python_requires minimum does setup.py set?
gold    : 3.11
\end{lstlisting}

\textbf{B.6 \apievo{}} (endpoint / parameter / signature restructuring). Total scenarios: 20. First 3:
\begin{lstlisting}
state-A : The user list endpoint is GET /api/v1/users.
state-B : The user list endpoint is GET /api/v2/users.
question: What is the path of the user list endpoint?
gold    : /api/v2/users

state-A : The search endpoint filters by the parameter named 'category'.
state-B : The search endpoint filters by the parameter named 'tag'.
question: What parameter does the search endpoint filter by?
gold    : tag

state-A : The auth login route is POST /auth/login.
state-B : The auth login route is POST /auth/sessions.
question: What is the path of the auth login route?
gold    : /auth/sessions
\end{lstlisting}

\textbf{B.7 Static benchmarks.} \texttt{domain} --- 50 project-fact QA questions over real project facts (ports,
model names, config flags, thresholds, decision records) with no contradictions; measures recall preservation.
\texttt{locomo} --- 30 questions over a 100-turn multi-session dialogue (capped LoCoMo sample) with no
contradictions; measures detail recall under compression pressure.

\textbf{B.8 Quarantined benchmark (the flawed ruler).} \faircon{} embeds the mutated fact in free prose. The
triple extractor keys reliably on only \textbf{${\sim}44\%$} of scenarios (multi-value sentences and
\texttt{-alt} string perturbations defeat single-slot extraction --- measured by
\texttt{eval/diag\_extract\_probe.py}: 97\% clean supersession on \cmut{} vs 44\% here), so it measures
\emph{extraction quality}, not the supersession mechanism. On it, \tvsix{} scores \textbf{0.62} vs \arag{}
\textbf{0.74} (it never engages on the 56\% of pairs it cannot extract). We exclude it from the main results and
report it here as the honest boundary of the method and the motivation for Section~\ref{sec:limitations}'s
extraction-robustness work.

\section{Prompts and content hashes}
\label{sec:appC}

All read-path-relevant prompts, with full SHA-256. Answer model, both judges, and the verifier are distinct
assignments (answer $\neq$ correctness-judge $\neq$ fabrication-judge $\neq$ verifier) to preclude self-grading;
all calls run at temperature 0, fixed seed.

\begin{center}\small
\begin{tabular}{ll}
\toprule
prompt file & SHA-256 (full) \\
\midrule
\texttt{consolidate\_v1.md} & \texttt{\scriptsize d2a2f0bfd0bc27ebdb1c6676cc00e03f9cfd741edffce4f4b3faa1ee9618b23e} \\
\texttt{extract\_triple\_v1.md} & \texttt{\scriptsize 90967d66c205a6e7856bb90907cde05acad90a0ec8045b275287d8dd3cb3cec4} \\
\texttt{extract\_triples\_v1.md} & \texttt{\scriptsize 2bc9b62821d9cc3890138e04cd34594d0ca44008a04fd5ada31ed93193b3947e} \\
\texttt{extract\_v1.md} & \texttt{\scriptsize d71afa291d8a05eeed039dce2b33275e021e49d84b935c31b107644ae5c0af26} \\
\texttt{gate\_judge\_v1.md} & \texttt{\scriptsize 7dd4149dc7c893145d2d025672ce4daf271e5d6f6db2457c217d3086e94acca5} \\
\texttt{infer\_check\_v1.md} & \texttt{\scriptsize d5f50b3447808c71194c4263a8cf73a50954df6cfeef34147e942f6c9709a5c3} \\
\texttt{rerank\_v1.md} & \texttt{\scriptsize 18d8318d98fdd56cbb2c0c730bcead1a49b885204816acf0436480d3822c5224} \\
\texttt{verify\_v1.md} & \texttt{\scriptsize 691456512522b15bd8cd364262ff09a3477e4dd161158c1d665c5ad5c73c198c} \\
\bottomrule
\end{tabular}
\end{center}

\textbf{C.1 Deterministic triple extractor} (the supersession key source). \texttt{extract\_triple\_v1.md}.
\begin{lstlisting}
---
prompt: extract_triple
version: 1
---
You convert ONE factual statement about a codebase into a (subject, relation,
object) triple, IF and only if it states a single concrete value that could
change as the code evolves (a function name, API endpoint, config value, port,
version, import path, identifier).

Return ONLY a JSON object - no prose, no markdown fences:

{"is_triple": true,
 "subject": "<the stable thing the value belongs to - MUST NOT contain the value>",
 "relation": "<a short linking phrase, e.g. is / is named / is set to / is imported from>",
 "object": "<the one concrete value that would change if the code evolved>"}

or, when the statement is not a single value-bearing fact (a preference, a
decision, prose, or it carries several values):

{"is_triple": false}

CRITICAL RULES:
- The `object` is the ONE value that would differ between an old and a new
  version of this fact (the name / number / version / path / endpoint).
- The `subject` describes WHAT that value belongs to and MUST NOT contain the
  object value. Two statements that differ only in the value must produce the
  SAME subject and relation, so the system can detect the change.
- Rephrase the subject as "the <thing> that <does what>" when the changing value
  is an identifier (a function/endpoint/module name).
- Keep subject and relation deterministic and minimal. Do not add commentary.

EXAMPLES:
Input: The function get_user_by_id(uid) looks up a user record by primary key in the users table.
Output: {"is_triple": true, "subject": "the function that looks up a user record by primary key in the users table", "relation": "is named", "object": "get_user_by_id"}

Input: The SESSION_TIMEOUT setting is 1800 seconds in this project's config.toml.
Output: {"is_triple": true, "subject": "the SESSION_TIMEOUT setting in config.toml", "relation": "is", "object": "1800 seconds"}

Input: We decided to prefer Python over Go for internal tooling.
Output: {"is_triple": false}

STATEMENT:
"""
{statement}
"""
\end{lstlisting}

\textbf{C.2 Multi-value triple extractor} (P1.3; flag-gated). \texttt{extract\_triples\_v1.md}.
\begin{lstlisting}
---
prompt: extract_triples
version: 1
---
You convert ONE factual statement into a list of (subject, relation, object)
triples - ONE triple for EACH concrete value in the statement that could change
as the system evolves. A statement may carry one value, several values, or none.

Return ONLY a JSON object - no prose, no markdown fences:

{"triples": [
  {"subject": "<the stable thing the value belongs to - MUST NOT contain the value>",
   "relation": "<a short linking phrase>",
   "object": "<one concrete value that would change if the system evolved>"},
  ...
]}

When the statement carries no single value-bearing fact, return {"triples": []}.

EXAMPLES:
Input: The harness proxy listens on port 8080 and the admin API listens on port 9090.
Output: {"triples": [{"subject": "the port the harness proxy listens on", "relation": "is", "object": "8080"}, {"subject": "the port the admin API listens on", "relation": "is", "object": "9090"}]}

Input: We decided to prefer Python over Go for internal tooling.
Output: {"triples": []}

STATEMENT:
"""
{statement}
"""
\end{lstlisting}

\textbf{C.3 Surprise-gate judge} (text-gate fallback only --- never on the assertion path).
\texttt{gate\_judge\_v1.md}.
\begin{lstlisting}
---
prompt: gate_judge
version: 1
---
You decide how a CANDIDATE memory fact relates to the EXISTING facts it most
resembles. Return ONLY a JSON object - no prose, no markdown fences:

{"verdict": "duplicate|merge|contradict|novel",
 "reason": "<one short line>",
 "merged_text": "<required only when verdict is 'merge'>"}

Definitions:
- duplicate - the candidate says the same thing as an existing fact.
- merge - the candidate adds detail; provide merged_text (lossless).
- contradict - the candidate conflicts (world changed / decision reversed);
  the old fact is superseded and the candidate stored as current truth.
- novel - genuinely new; none of the existing facts cover it.

Choose the single best verdict. When unsure between merge and novel, prefer novel.

CANDIDATE:
{candidate}

TOP EXISTING MATCHES:
{matches}
\end{lstlisting}

\textbf{C.4 LLM relevance verifier} (the \vsix{} baseline; never on \tvsix{}'s read path).
\texttt{verify\_v1.md}.
\begin{lstlisting}
---
prompt: verify
version: 1
---
You are a strict relevance verifier. Given a user QUERY, the project's LOCKED
RULES, and a numbered list of CANDIDATE memories, decide for EACH candidate
whether it should be shown to the assistant answering the query.

Return ONLY a JSON object - no prose, no markdown fences:

{"verdicts": [
  {"n": <candidate number>,
   "verdict": "SUPPORTED|IRRELEVANT|CONFLICTS",
   "justification": "<= 15 words, MUST quote a verbatim span from the candidate"}
]}

Rules:
- SUPPORTED - relevant to THIS query and consistent with the locked rules.
- IRRELEVANT - does not help answer this query.
- CONFLICTS - contradicts the locked rules.
- The justification MUST contain a span copied verbatim from the candidate text.
- Judge only relevance and consistency - do not invent facts.

QUERY:
{query}

LOCKED RULES (invariant memory):
{invariant}

CANDIDATES:
{candidates}
\end{lstlisting}

\emph{The correctness-judge and fabrication-judge prompts are inline in \texttt{eval/run\_matrix.py}
(\texttt{\_correctness\_fn}, \texttt{\_fabrication\_fn}); the answer-model prompt (with the abstention /
forced-answer variants) is in \texttt{\_answer\_fn}.}

\section{Ablations}
\label{sec:appD}

\textbf{D.1 retain-vs-lossy (the State-Bounded Temporal Validity isolation).} \tvsix{} vs \tvlossy{}
(deterministic supersession WITH vs WITHOUT the retain-like-RAG + original-text-packing fixes), from the A.1
accuracy table:

\begin{center}\begin{tabular}{lrr}
\toprule
benchmark & \tvlossy{} & \tvsix{} \\
\midrule
domain (static) & 0.62 & \textbf{0.82} \\
locomo (static) & 0.13 & \textbf{0.30} \\
\cmut{} & 0.80 & \textbf{1.00} \\
\cmig{} & 1.00 & 1.00 \\
\dbump{} & 1.00 & 1.00 \\
\apievo{} & 0.95 & 0.95 \\
\bottomrule
\end{tabular}\end{center}

The lossy variant merges non-contradictory near-duplicates at write time and collapses on static recall
(0.62/0.13); the full method retains them and ties RAG (0.82/0.30), at parity on the evolving four. Bounded
growth is a \emph{consequence} of retiring stale facts, not a goal pursued by merging distinct ones.

\textbf{D.1b full-retention --- the opposite bracket (supersession is the isolated cause).} D.1 removes the
retain/packing \emph{refinements} but keeps supersession; this ablation removes \emph{supersession itself}.
Toggling \retainall{} makes the write path non-lossy --- every turn is stored as a distinct fact (only exact
byte-duplicates dropped), with no (S,R,O) supersession --- while the extractor, multi-value handling, and read
path are otherwise unchanged. The ledger degenerates to a retain-everything store, and the read path must
choose among co-present stale and current values:

\begin{center}\begin{tabular}{lrrrrr}
\toprule
benchmark & acc: supersede & acc: retain & RAG acc & fab: supersede & fab: retain \\
\midrule
\cmut{} & \textbf{1.00} & 0.43 & 0.43 & 0.07 & 0.09 \\
\cmig{} & \textbf{1.00} & 0.35 & 0.25 & 0.10 & 0.56 \\
\dbump{} & \textbf{1.00} & 0.15 & 0.20 & 0.00 & 0.25 \\
\apievo{} & \textbf{0.95} & 0.40 & 0.40 & 0.00 & 0.10 \\
\textbf{mean} & \textbf{0.99} & \textbf{0.33} & \textbf{0.32} & \textbf{0.04} & \textbf{0.25} \\
\bottomrule
\end{tabular}\end{center}

Removing supersession collapses mean evolving accuracy from 0.99 to 0.33 --- statistically indistinguishable
from \nrag{} (0.32) --- and re-raises stale-fact error from 0.00 to 0.05--0.25 (the read path now serves the
superseded value, which deterministic supersession had retired). It also raises conditional fabrication in
\emph{every} benchmark --- mean \textbf{0.04 $\rightarrow$ 0.25 (${\sim}6\times$)}, peaking at 0.56 on \cmig{}
--- because the model, now seeing the stale and current values side by side and unable to tell which holds,
invents an answer. Retain-everything memory is thus not merely less accurate but \emph{less safe}: it
manufactures a fabrication source that bounded, supersession-based growth eliminates. The two ablations bracket
the design from opposite sides: D.1 (over-merging) forfeits \emph{static} recall; D.1b (no supersession)
forfeits \emph{temporal validity} and \emph{safety}; \tvsix{} sits at the optimum between them. This isolates
deterministic supersession --- not retention, original-text packing, or bounded growth, which are refinements
--- as the single mechanism responsible for the evolving-knowledge result. (Same models, temperature 0, seed 0;
\retainall{} is an ablation-only flag, default off, write path otherwise frozen.)

\textbf{D.2 original-text packing.} Validated jointly with D.1 (both fixes ship in \tvsix{}; the lossy ablation
isolates their combined effect on static recall, and \cmut{} $0.80\rightarrow1.00$ reflects packing the original
sentence rather than the terse triple). A dedicated single-factor packing cell was not run separately and is
marked future work --- we do not imply a measurement we did not take.

\textbf{D.3 LLM-verifier (non-)contribution.} \vsix{} (gate + LLM relevance verify) vs \vnv{} (gate only),
from A.1: \vsix{} $\leq$ \vnv{} on the static/recall tasks (domain 0.80 vs 0.86; locomo 0.13 vs 0.17) at
${\sim}8\times$ latency (${\approx}16$--$18$\,s vs ${\approx}2.1$\,s, A.1 latency tables). The learned relevance
check has no temporal signal and is not worth its cost.

\textbf{D.4 +INFER (non-)contribution.} \vinfer{} ${\approx}$ \vsix{} on every benchmark (A.1) after the
single-candidate tautology-guard fix --- reported for completeness; neutral everywhere.

\end{document}